\documentclass[letterpaper, 10 pt, conference]{ieeeconf}
\IEEEoverridecommandlockouts
\usepackage{cite}
\usepackage{amsmath,amssymb,amsfonts}
\usepackage{xfrac}
\usepackage{algorithmic}
\usepackage{graphicx}
\usepackage{textcomp}
\usepackage{xcolor}
\def\BibTeX{{\rm B\kern-.05em{\sc i\kern-.025em b}\kern-.08em
    T\kern-.1667em\lower.7ex\hbox{E}\kern-.125emX}}

\usepackage[capitalize]{cleveref} 

\title{\LARGE \bf Long-Reach Robotic Manipulation for \\ Assembly and Outfitting of Lunar Structures}

\author{Stanley Wang*, Venny Kojouharov*, Long Yin Chung, Daniel Morton, and Mark Cutkosky%
\thanks{* Equal contribution.}%
\thanks{All authors are with the Dept. of Mechanical Engineering, Stanford University, Stanford, CA 94305, USA.}%
\thanks{Corresponding author email: {\tt\small swang11@stanford.edu}}}

\begin{document}

\maketitle

\begin{abstract}
Future infrastructure construction on the lunar surface will require semi- or fully-autonomous operation from robots deployed at the build site. In particular, tasks such as electrical outfitting necessitate transport, routing, and fine manipulation of cables across large structures. 
To address this need, 
we present a compact and long-reach manipulator incorporating a deployable composite boom, capable of performing manipulation tasks across large structures and workspaces. We characterize the deflection, vibration, and blossoming characteristics inherent to the deployable structure, and present a manipulation control strategy to mitigate these effects. Experiments indicate an average endpoint accuracy error of less than 15 mm for boom lengths up to 1.8\,m. We demonstrate the approach with a cable routing task to illustrate the potential for lunar outfitting applications that benefit from long reach.

\end{abstract}

\begin{keywords}
Space robotics, manipulation, deployable, in-space manufacturing and assembly
\end{keywords}

\section{Introduction}
%
The dawn of commercial lunar activities (e.g. NASA CLPS \cite{nasa_clps_2025}) 
brings many exciting opportunities and challenges in the assembly and construction of surface and subsurface infrastructure \cite{arney2024space,belvin2016space}. 
Examples include the preparation of sites for extended operation and eventual long-term human habitation \cite{benaroya2017lunar}. Such structures will often depend critically on either fabrication via available on-site materials (In-Situ Resource Utilization) \cite{ellery2020sustainable} or adaptation of existing subsurface geology (e.g. lava tubes) \cite{coombs1992search}. 
In either case, they will require significant secondary construction and outfitting to become serviceable. Robots and autonomous systems will undoubtedly serve a critical role in augmenting this process, assisting in tasks such as installing cables and tubing, affixing sensors and components, drilling holes, etc. For overviews, see \cite{arney2024space,belvin2016space}.



\begin{figure}
\centering
\includegraphics[width=0.95\columnwidth]{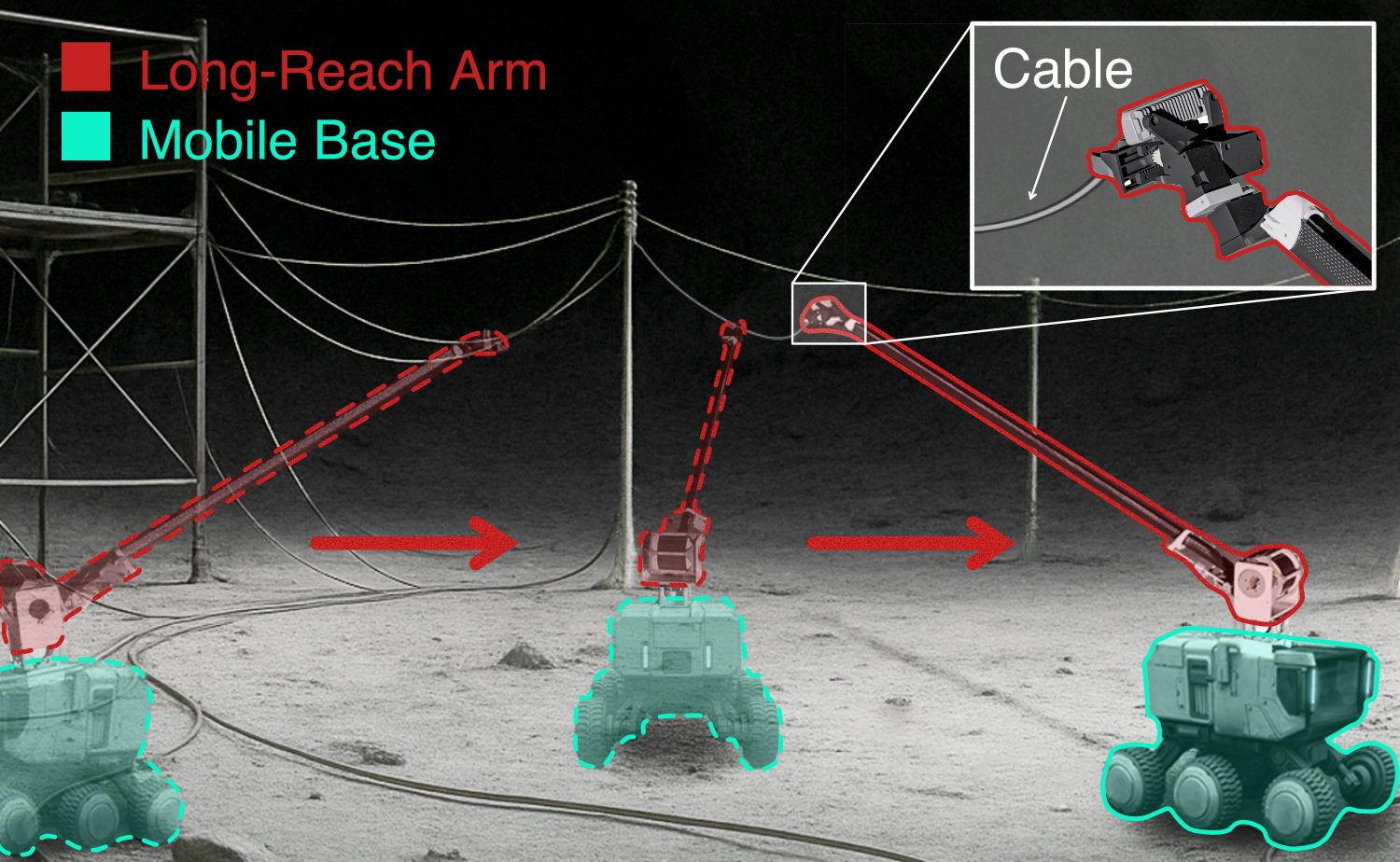}
\caption{Small mobile robots (green) equipped with long, extendable booms (red) can perform tasks that require a large workspace---such as cable stringing and assembly for lunar construction.}
\label{fig:concept}
\end{figure}

We propose that a small robot with extensive reach, as shown in \cref{fig:concept}, will be highly applicable to tasks in this domain. Depending on the application, the base could be a wheeled or tracked rover (e.g. \cite{lehner2018mobile, martinez2023multi}) or a quadruped (e.g. \cite{richter2024multi,kolvenbach2021quadrupedal}). The focus in this paper is on the addition of an arm and wrist, with a very long reach provided by a deployable composite boom. This combination enables operation over a workspace of 10x-20x the length scale of the mobile base. It allows a light, compact, and easily transported rover to perform tasks that would otherwise require a larger and heavier robot with conventional jointed arms.

Although the extending boom offers clear advantages in workspace and mass, it also presents challenges in force application and precise positioning. The boom is harder to control during extension and retraction than a conventional prismatic axis and has lower bending and torsional stiffness, leading to elastic deflections and structural vibrations.

This work serves as a preliminary investigation into the application of long-reach structures for dexterous lunar manipulation. In the following sections, we review related work, introduce our boom deployment and positioning mechanism, and analyze key effects such as boom deflection and “blossoming’’ that occurs when wrapping a composite boom around a drum. We present strategies for mitigating these effects through modeling, visual endpoint servoing, and pose-dependent velocity adjustment, and demonstrate their effectiveness in a three-dimensional cable-stringing task. Finally, we discuss future directions, including longer booms scaled to lunar gravity and mission architectures with coordinated mobile bases.

\begin{figure*}[t]
    \vspace{5pt}
    \centering
    \includegraphics[width=\textwidth]{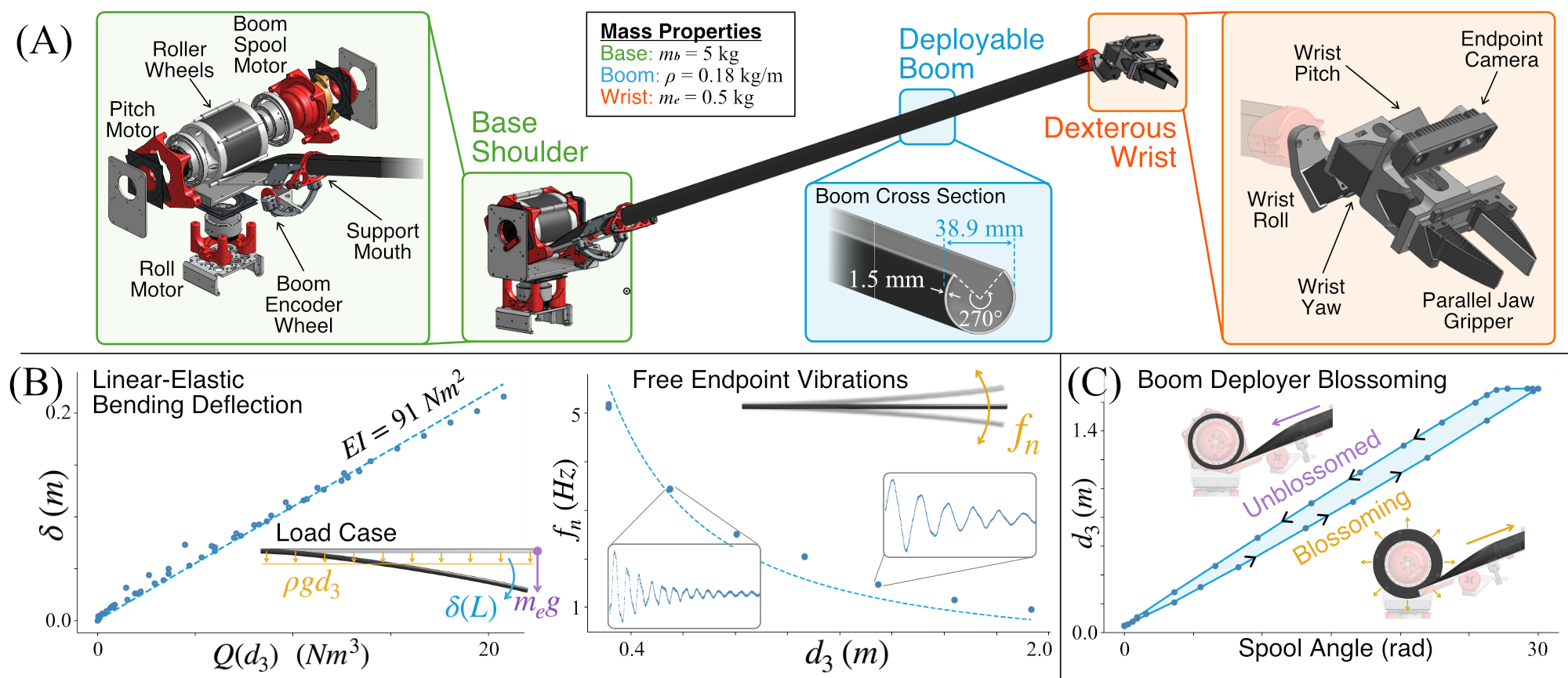}
    \caption{
    (A) Overview of key hardware components in the long-reach manipulator design: shoulder base, deployable boom arm (0.2 - 3m), and dexterous wrist. 
    (B) Mechanical characterizations of the boom's uniaxial bending and free vibration effects. (C) Boom deployment blossoming effects.
    }

    \label{fig:hardware_design}
\end{figure*}

\section{Related Work}


We draw inspiration from numerous developments in deployable technologies.
Deployable space booms utilizing advanced composite materials play a critical role in spacecraft structures, including antennae, deorbit sails \cite{stohlman2013deorbitsail}, and solar arrays \cite{straubel2011deployable}.
Notable examples include CTM (Collapsible Tube Mast) \cite{fernandez2018bistable}, STEM (Storable Tubular Extendible Member) \cite{Thomson1993DeployableAR}, and TRAC (Triangular Rollable and Collapsible) \cite{murphey2017trac}.

Roboticists have also recognized the utility of deployable structures in enabling lightweight long-reach field manipulation. An early example was the metal arm for excavating regolith on the Viking program rover \cite{Viking}. More recent examples include booms, masts, and everting vine robots intended for manipulation \cite{collins2016design, zippermast2015, blumenschein2020design}. ReachBot \cite{chen2024locomotion} proposed several booms, similar to the ones employed here, to locomote in extreme terrain such as Martian caves. However, that work did not fully address hardware control limitations for the boom itself or demonstrations of dexterous manipulation, which are focuses of this study. 


Considering the particular task of cable routing for outfitting lunar structures, we draw upon extensive literature in modeling and manipulating deformable linear objects (DLOs), as reviewed in \cite{saha2007manipulation}. A lunar-specific approach to DLO modeling is further presented in \cite{quartaro2024modeling}. While such works have primarily been concerned with object-centric modeling and robot perception/control, the potential utility of a long-reach manipulator architecture has not been addressed---a gap that this work aims to fill.






\section{Hardware Systems}
We present several key hardware contributions in leveraging deployable booms as long-reach manipulator arms, as shown in \cref{fig:hardware_design}. Through empirical modeling, we inform design improvements to the boom mechanisms, while full workspace dexterity is enabled by a shoulder base combined with a three‑axis wrist gripper.



\subsection{Fiberglass Boom Characterization}
The deployable actuator of our system utilizes a 3\,m rollable fiberglass boom from Metolius Climbing \cite{metolius2025rollup}. We find this off-the-shelf component to offer a good compromise between performance and affordability $(<\$300)$.
The boom’s geometry resembles a slit-tube channel, roughly modeled as a 270$^\circ$ circular sector with diameter 38.9\,mm and wall thickness 
1.5\,mm. This open-channel beam is relatively weak in torsion, but maintains acceptable bending stiffness while being exceptionally easy to unroll and deploy. The matrix material of our fiberglass boom is also quite resilient, and can withstand hundreds of deployment cycles before fatigue degradation.

Considering this deployable as a prismatic (extension) joint for robotic manipulation, we find it critical to characterize the boom’s non-rigid effects, specifically elastic deformations and vibrations (Fig. \ref{fig:hardware_design}B) and how they scale with boom length $(d_3)$. By orienting the strongest bending axis of our boom along the (dominant) gravitational loads, we consider a simplified uniaxial bending model. Empirical results in Fig. \ref{fig:hardware_design}B show behavior consistent with Euler-Bernoulli theory, with linear-elastic deflection against generalized load factors $Q(d_3)$:
\begin{equation}
    \delta(d_3) = \frac{1}{EI}\Big(\underbrace{\frac{1}{3}m_e gd_3^3 + \frac{1}{8}\rho g d_3^4}_{=Q(d_3)}\Big)
\end{equation}
where $m_e$ is the mass at the boom's endpoint, $d_3$ is the boom length, $E$ is the composite Young's modulus (ignoring anisotropicity), $I$ is the cross-sectional area moment, $\rho$ is the boom linear density, and $g$ is the local gravitational constant.

From our data, we extract a lumped flexural rigidity (EI) indicative of the general compliance of our beam against gravitational loads.


Free vibrations of the mechanical structure are also studied by exciting a small angular impulse about the boom’s base and measuring resultant endpoint acceleration. The natural frequencies and damping are estimated by fitting damped sinusoids, and plotted in Fig. \ref{fig:hardware_design}B across different lengths.

\subsection{Boom Deployer Design}

Deployment of the boom is controlled by a motorized spool and radial support rollers, wherein rotation produces either an unspooling (extension) or spooling (retraction) at up to $0.4 m/s$.
Blossoming is a notable issue in this mechanism, where the spooled boom may partially uncoil or expand during deployment.
As shown in Fig. \ref{fig:hardware_design}C, this leads to uncertainty between the spool’s controlled rotational position and the actual deployed length of the boom. While mechanical countermeasures exist, such as constraining the spool between spring-loaded rollers or co-winding with a secondary tape \cite{firth2019deployment}, these approaches introduce additional friction and complexity and reduce reliability for robotic applications. We instead overcame this issue by attaching an encoder wheel directly against the boom. This provides a highly accurate measurement of boom velocity, allowing closed-loop control compensation. 

\subsection{Robot Manipulator Design}
For our overall manipulator as shown in Fig. \ref{fig:hardware_design}A, we adopt a macro–mini inspired architecture \cite{khatibmacro} comprising a base for large-scale reorientation $(\theta_1, \theta_2)$, the deployable boom for extended reach $(d_3)$, and a wrist for fine manipulation $(\theta_4,\theta_5,\theta_6)$. The base employs two high-torque brushless planetary gear motors (CubeMars T-Motor AK60-36) to control the pitch and roll of the boom. At the boom’s endpoint, three Dynamixel XL430-W250-T servomotors are arranged with intersecting axes to form a compact wrist manipulator. Finally, a simple parallel-jaw gripper equipped with compliant fingers $(m_e = 0.5 \,kg)$ provides grasping capability. The entire manipulator assembly is under 6\,kg, and can support additional endpoint payloads over $250\,g$.


\section{Control Framework}

We implement a hybrid control architecture (\cref{fig:control}), combining joint velocity control, a PI feedback loop for boom deployment, deflection-compensated kinematics, and vision-based endpoint servoing to track task trajectories. Real-time perception and control were executed on a Jetson AGX Orin.

\subsection{Joint Control}
All five revolute joints $(\theta_i)$ operate with position-based velocity control at 100 Hz. The discrete-time update law is:
\begin{equation}
    \theta_{t+1} = \theta_t + \Delta t \; \dot\theta_{t,ref}
\end{equation}
This approach allows us to easily set motor characteristics (i.e. joint limits and impedances) to resolve safe, smooth joint trajectories. 
For the deployable prismatic joint $(d_3)$, we design our own PI feedback loop to stabilize the commanded joint velocity $(\dot d_3)$
\begin{equation}
\dot d_3 = \dot d_{3,\text{ref}} + K_p(\dot d_{3,\text{ref}} - \dot d_{3,o}) + K_I\int(\dot d_{3,\text{ref}} - \dot d_{3,o}) dt
\end{equation}
where $\dot d_{3,\text{ref}}$ is the desired velocity and $\dot d_{3,o}$ 
is the actual boom velocity as measured by the encoder wheel. When blossoming occurs, integral error accumulation quickly increases actuator effort, enabling the controller to overcome the effect.

\subsection{Robot Kinematics}

\begin{figure}[t]
\vspace{5pt}
\centering
\includegraphics[width=0.95\columnwidth]{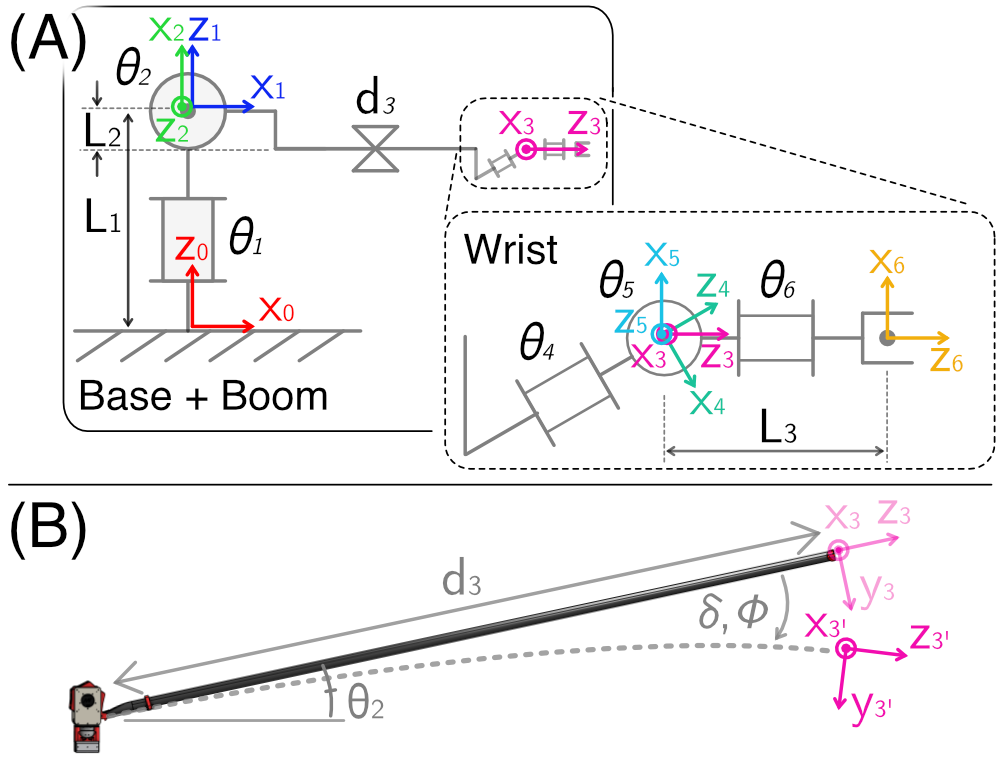}
\caption{(A) Rigid kinematic model of the robot manipulator following an R-R-P-R-R-R architecture. (B) Additional augmentation transformation used to compensate for elastic boom deflection. 
}
\label{fig:kinematics}
\end{figure}

A conventional rigid-body model is first used to establish an approximation of our robot kinematics, as shown in Fig. \ref{fig:kinematics}. Our six generalized joint coordinates consist of:
\begin{equation}
    \mathbf{q} =\big[
\underbrace{\theta_1, \theta_2}_{\text{base}},
\underbrace{d_3}_{\text{boom}},
\underbrace{\theta_4, \theta_5, \theta_6}_{\text{wrist}}
\big]
\end{equation}
We resolve the forward kinematics of the end-effector as:
\begin{equation}
T_0^{6} = T(\mathbf{q}) = T_0^{1} T_1^{2}\ldots
\end{equation}
However, we additionally augment the boom tip (after $d_3$) with a transform from our linear-elastic deflection model. The total linear deflection $(\delta)$ and rotational deflection $(\phi)$ from our load case are given as:
\begin{align}
    \delta(\mathbf{q}) &= \frac{1}{EI}\cos\theta_2 \left(\frac{1}{3}Mgd_3^3 + \frac{1}{8}\rho g d_3^4\right)\\
    \phi(\mathbf{q}) &= \frac{1}{EI}\cos\theta_2\left(\frac{1}{2}Mgd_3^2 + \frac{1}{6}\rho g d_3^3\right)
\end{align}
These compose our deflection augmentation transform between frame $\{3\}$ and frame $\{3'\}$:
\begin{equation}
T_{def} = T_3^{3'} = \begin{bmatrix}
    R_{x3}(-\phi)  & \delta\;\mathbf{\hat{y}_3} \\
    0 & 1
\end{bmatrix}
\end{equation}
Corrected forward kinematics with boom deflection become:
\begin{equation}
T'(\mathbf{q}) = T_0^3\;T_{def}\;T_{3'}^6
\end{equation}
We found this model to be representative of quasistatic behavior, where relatively slow motions of the boom endpoint avoid exciting significant inertial or resonant dynamics.

The linear and angular velocity Jacobians are computed from the forward kinematics (including deflection compensation) such that the end-effector twist is given by:
\begin{align}
\mathbf{v} = \begin{bmatrix}
    v \\ \mathbf{\omega}
\end{bmatrix}
= \begin{bmatrix}
    J_v(\mathbf{q})\\
    J_\omega(\mathbf{q})
\end{bmatrix}
\dot{\mathbf{q}}
\end{align}
Using the pseudoinverse of the Jacobian (to avoid instabilities near singularity) allowed us to relate task-space velocities $\mathbf{v}$ back to joint space $\mathbf{\dot q}$.
\begin{equation}
\mathbf{\dot q} = J^\dag(\mathbf{q})\mathbf{v} \qquad J^\dag = J^T(JJ^T)^{-1}
\end{equation}

\subsection{Endpoint Visual Servo}
To overcome the model-based shortcomings in control, we resolve a closed-loop visual feedback system for task-oriented positional accuracy.
We use only onboard sensing for our autonomy stack (i.e., no external motion capture feedback), which makes the design more generalizable and better suited for field deployment.
To this end, a single calibrated RGB camera (Luxonis Oak-D S2) is mounted distally at the gripper, allowing for endpoint perception. AprilTag 36h11 markers \cite{olson2011apriltag} are then arranged around the space of interest, such that we resolve the pose of the end-effector relative to the target. Multiple tags combined with temporal filtering of poses (via SLERP \cite{kremer2008quaternions}) were used to avoid occlusion and ensure tracking continuity.

\begin{figure}[t]
\vspace{5pt}
\centering
\includegraphics[width=\columnwidth]{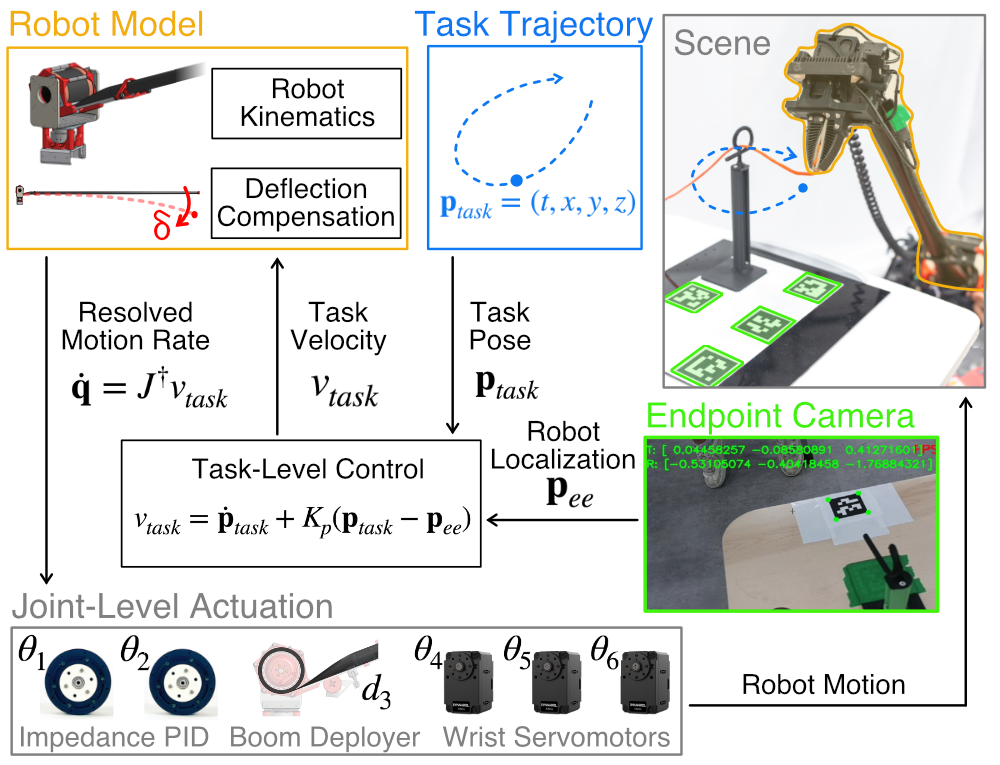}
\caption{Overview of the control framework. The robot follows task trajectories using visual end-effector localization and an augmented task-space kinematic model for closed-loop position control.}
\label{fig:control}
\end{figure}

\subsection{Task-Space Motion Control}
\Cref{fig:control} provides an overview of our closed-loop control stack. 
To execute a manipulation task, we specified a desired end-effector trajectory in the frame of the  task target (i.e. hook fixtures for cable routing)
$\mathbf{p_{task}} = (x,y,z,t)$.
The wrist-mounted camera then detected AprilTags affixed to the target and provided an estimate of end-effector pose $\mathbf{p_{ee}}$.
A proportional feedback law with velocity feedforward generated the commanded task-space velocity as:
\begin{equation}
\mathbf{v_{task}} = \dot{\mathbf{p_{task}}} + K_p\left(\mathbf{p_{task} - p_{ee}}\right)
\end{equation}
Desired joint velocities were then obtained using resolved rate control $\dot{\mathbf{q}} = J^\dag(\mathbf{q}) \mathbf{v_{task}}$.


\begin{figure*}[t]
    \vspace{5pt}
    \centering
    \includegraphics[width=0.95\textwidth]{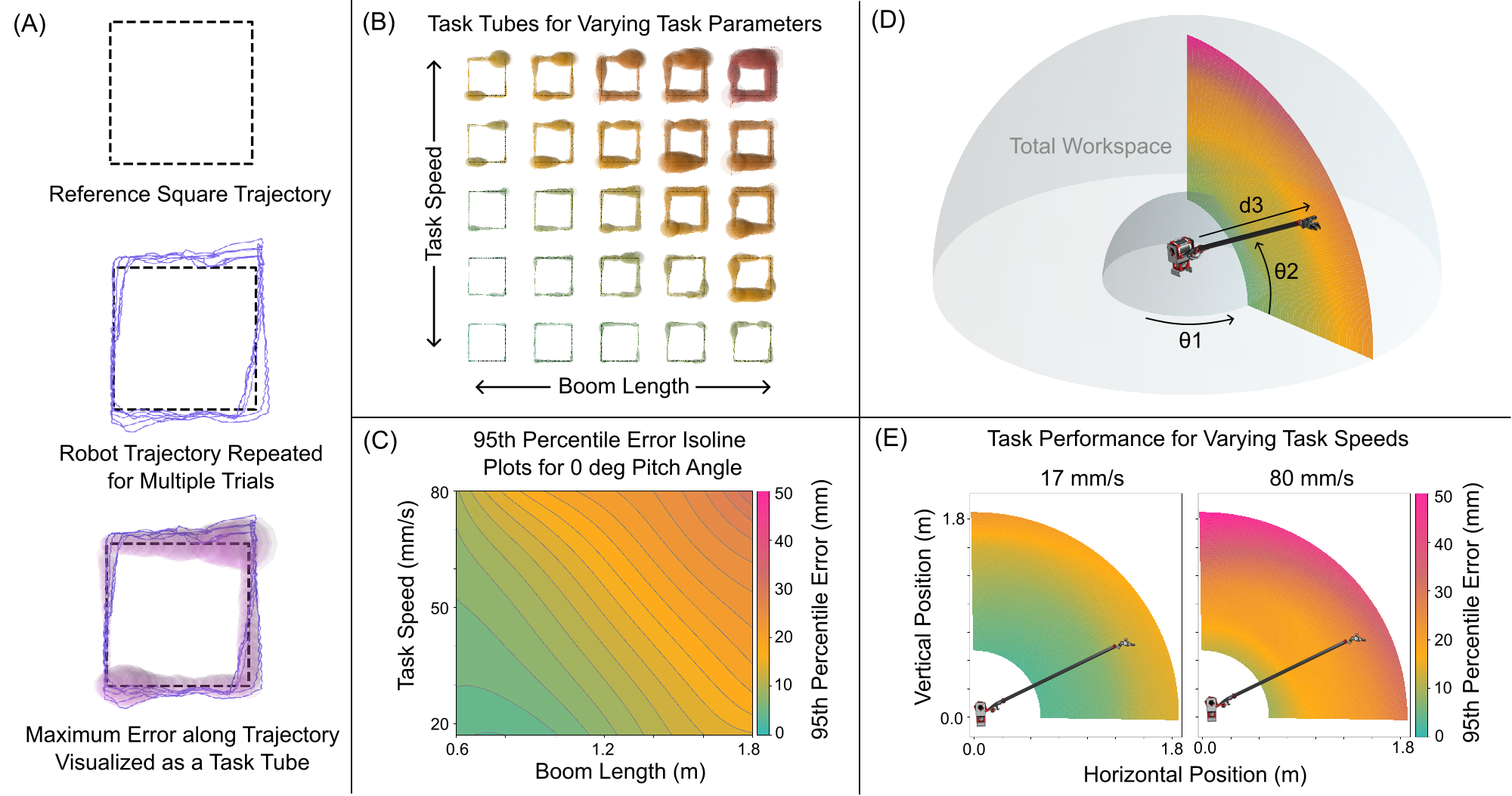}
    \caption{%
        Summary of tracking error across the robot workspace.
        (A) A $10\,\text{cm} \times 10\,\text{cm}$ square reference trajectory. The robot’s executed trajectories across trials are overlaid, and the maximum spatial error is visualized as a task tube.
        (B) All trajectories executed at $\theta = 0^\circ$ for varying boom lengths (horizontal axis) and task speeds (vertical axis). Error increases with both parameters.
        (C) Interpolated 95th percentile isoline plot for $\theta = 0^\circ$. 
        (D) Total robot workspace schematic showing boom length $d_3$, pitch angle $\theta_2$, and yaw $\theta_1$ with an error slice at 50\,mm/s. 
        (E) Interpolated error slices at $\theta \in [0^\circ, 90^\circ]$ for two speeds: 17\,mm/s (slow) and 80\,mm/s (fast). Errors increase with both pitch angle and extension, especially at higher speeds.%
    }
    \label{fig:experiments}
\end{figure*}

\section{Experimental Robot Evaluation}

We experimentally benchmarked the robot's accuracy in executing a reference trajectory using three robot parameters that encompass the robot's workspace: pitch angle ($\theta_2$), boom length ($d_3$), and task speed ($s$).

\subsection{Task Error Metric}

To quantify accuracy, we define a ``task tube" as a spatial envelope that bounds the error of the robot trajectory relative to a reference trajectory as seen in \cref{fig:experiments}A. This concept is inspired by tracking error tubes used in motion planning frameworks \cite{compton2024dynamic}.

To construct a task tube, a reference trajectory is discretized into a set of goal points $\{g_i\}$, and for each goal point, we find all actual robot trajectory points $\{r_j\}$ for which $g_i$ is the nearest neighbor. Then we find the maximum distance $\|r_j - g_i\|$ at each goal point and use that to form a uniform sphere around the goal point. This process gives a spatial envelope that bounds the trajectory, capturing deviations in all directions.

The task tube provides an intuitive geometric visualization of error and can show how uniformly distributed the error is in space. Furthermore, we use the task tube to define task error, $e$, to be the 95th percentile of task tube radii distribution. We use the 95th percentile error rather than the maximum because it provides a reasonable estimate of task accuracy while minimizing the effect of outliers due to noise and other disturbances.


\subsection{Experimental Design and Data Collection}

For our experiments, we define the reference trajectory to be a 10\,$\times$10\,cm square trajectory. This trajectory requires comparable effort in both prismatic and revolute joints and includes sharp corners that can induce large tracking errors. It therefore yields conservative performance metrics when generalizing to arbitrary trajectories (e.g. routing a cable through a hook). 

Experiments were performed with variations in  three parameters. The pitch angle of the boom was varied across {$\theta_{2,i} \in \{0, 45, 90\}^\circ$}. The boom length was varied across {$d_{3j} \in \{0.6, 0.9, 1.2, 1.5, 1.8\}$}\,m and task speeds was varied across {$s_k \in \{17, 33, 50, 67, 80\}$}\,mm/s. Each configuration was repeated 5 times, yielding a total of 375 trials.

In each trial, the actual robot trajectory points $\{r_j\}$ were logged using an OptiTrack\texttrademark \,motion capture system. We plotted the task tube around the reference square trajectory in \cref{fig:experiments}B and recorded the task error, $e$.


\subsection{Interpolated Error Estimation}

To generalize the experimental results across the robot’s workspace, we constructed a spatially continuous approximation of task error, $e$, using trilinear interpolation:
\begin{equation}
e(\theta_2, d_3, s) = \text{Interp3D}(e_{ijk})
\end{equation}
where $e_{ijk} = e(\theta_{2,i}, d_{3,j}, s_k)$ are the experimentally measured tracking errors at discrete configurations.

This interpolated function provides an estimate of the expected task error for any combination of $(\theta_2, d_3, s)$ and serves as a compact, data-driven model of system performance that can be queried to inform configuration selection, trajectory generation, or speed adjustment for new tasks.

\subsection{Experimental Results}

\Cref{fig:experiments} summarizes the key results. \Cref{fig:experiments}B shows the task tubes of all trials collected at $\theta_2 = 0^\circ$, organized by boom length and task speed. Subsequently, \cref{fig:experiments}C shows an isoline plot of the interpolated error values, $e$, at $\theta_2 = 0^\circ$. Task error, $e$, increases monotonically with both parameters. 

\Cref{fig:experiments}D visualizes the task error,$e$, in a slice of the robot workspace, across different pitch angles and boom lengths. Lastly, \cref{fig:experiments}E show the error, $e$, in the same workspace slice for two representative speeds (17\,mm/s and 80\,mm/s). Error increased with pitch angle, with the lowest errors occurring when the boom is horizontal and the highest when vertical.

Across all configurations, the task error ($e$) ranges from approximately 3\,mm to 50\,mm, highlighting that pitch angle, boom length, and task speed are all significant contributors to end-effector accuracy.

\section{Application: Cable Routing Task}

As a proof of concept for lunar surface outfitting, we performed a semi-autonomous cable routing task using our long-reach manipulator arm. The robot was fixed to a static base and tasked with routing a cable through wall-mounted hooks placed at various orientations and locations.

\subsection{Human Routing Demonstration}
\begin{figure}[t]
    \vspace{5pt}
    \centering
    \includegraphics[width=0.9\columnwidth]{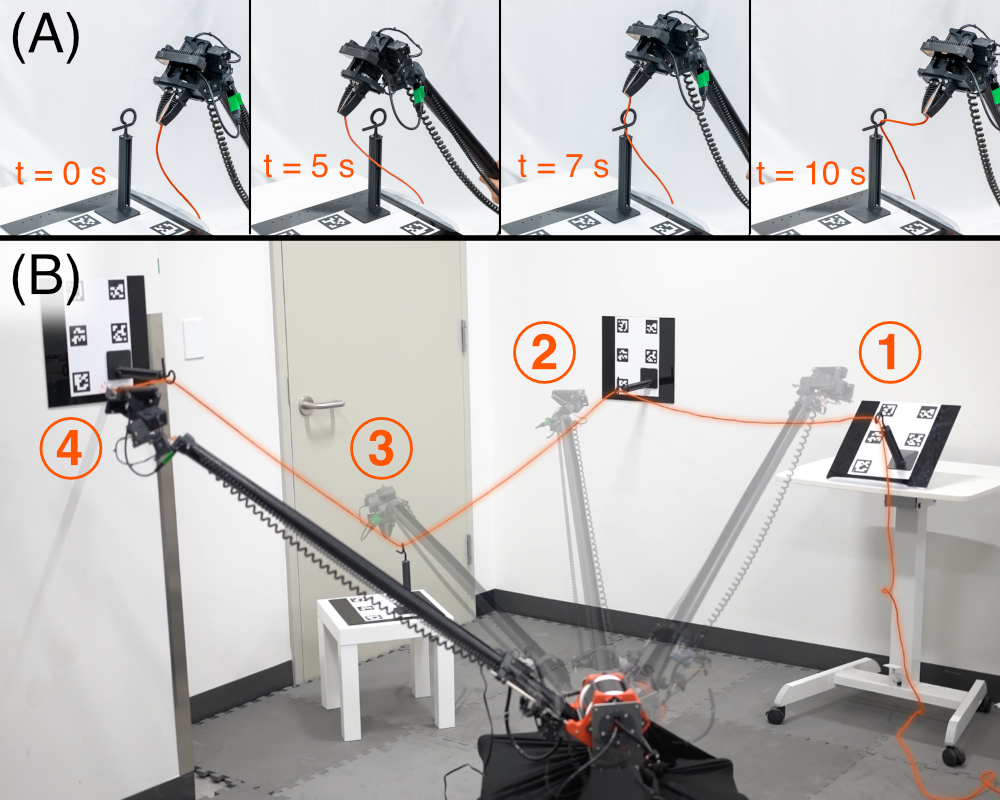}
    \caption{%
         Semi-autonomous cable routing task. (A) Four stages of the task motion primitive. (B) Robot autonomously performs this task primitive using visual servoing at four hooks at varying locations and orientations. \emph{Video available in supplementary material.} 
         %
    }
    \label{fig:cable demo}
\end{figure}

We used spiral hooks (e.g. Erico 4BRT20) for the cable routing demonstration (\cref{fig:cable demo}A) because, unlike closed loops, they allow the cable to be routed around--eliminating the need to insert or release the cable during routing. To generate the robot's reference cable routing trajectory, a human operator demonstrated a single hook routing motion using a tracked gripper proxy. This short motion was recorded and stored as a reusable primitive that the robot replays at different hook locations and orientations.

\subsection{Semi-Autonomous Cable Routing Task}

In the example task (\cref{fig:cable demo}B), four hooks marked with AprilTags were placed at varying orientations and locations in a
2m$\times$2m$\times$2m workspace. The robot started in a fully retracted configuration at one corner of the workspace. At each hook, it autonomously executed the cable routing primitive, while a human teleoperated it between hooks.

\subsection{Cable Routing Speed Selection}
Wtih a fixed robot base, the approximate boom length and pitch angle at each hook location are predetermined. To ensure reliable performance, the task speed $s^*$ was selected so that the task error, $e$, was below a threshold error, $\bar{e}$: 

\begin{equation}
e(\theta_2, d_3, s^*) \leq \bar{e}
\end{equation}

For our cable routing task we found $\bar{e} = 15$ mm to produce sufficient accuracy.

\begin{table}[ht]
\caption{Cable Routing Parameters and Errors at Each Hook}
\label{tab:hook_params}
\centering
\begin{tabular}{c|c|c|c|c|c}
\textbf{Hook} & ${\theta_2}$ ($^\circ$) & \textbf{$d_3$ (m)} & \textbf{$s^*$ (mm/s)} & \textbf{$e$ (mm)} & \textbf{$e_{meas}$ (mm)} \\
\hline
1 & 35 & 1.27 & 55 & 15 & 14.8 \\
2 & 24 & 1.78 & 19 & 15 & 11.4 \\
3 & 13 & 1.07 & 70 & 15 & 14.1 \\
4 & 31 & 1.55 & 25 & 15 & 10.9 \\
\end{tabular}
\end{table}

Using this approach, the robot successfully routed the cable through all four hooks without collision or failure as shown in \cref{fig:cable demo}. \Cref{tab:hook_params} summarizes the parameters and measured errors, $e_{meas}$, at each hook. In all cases, the measured error remained below the predicted task error, $e$, indicating that our experimentally derived error model provides a conservative estimate of real world task error.  


\section{Discussion and Future Work}

Our results demonstrate that long-reach manipulators with simple control strategies can effectively perform semi-autonomous manipulation tasks like cable routing. This section reflects on our experimental findings, structural and control limitations, and generalization to future lunar tasks and microgravity environments.

\subsection{Experimental Results Interpretation}

Our experiments show that tracking error increases with boom length, pitch angle, and task speed. Longer extensions amplify motion inaccuracies. There are structural issues due to elastic deflection, joint compliance, and vibrations. Also, as length increases, small angular deviations at the shoulder cause increasingly larger positional errors at the tip.

Regarding pitch angle, when the boom is horizontal or pointing downward, gravity tends to stabilize it from small disturbances. As the boom rotates upward toward vertical, gravitational forces are no longer restoring, making the system more subject to oscillations (like an inverted pendulum). In addition, faster motions exacerbate the unmodeled structural dynamics, producing greater lag and overshoot in control and decreasing task accuracy.

These relationships among robot configuration, speed, and task accuracy match the observed trends in our error data and motivate design, control, and modeling improvements, as discussed in the next section.

\subsection{Design Improvements}
Enhancements to the robot's mechanical structure can improve its manipulation capabilities.
Advancements in deployable technologies (e.g. stiffer composites resilient to high cyclic strain) will allow the arm to become longer, for larger workspaces--especially in reduced gravity. Stiffening the shoulder base is also important, as joint inaccuracies and compliance produce errors that grow linearly with boom length. The potential for larger endpoint forces and moments in other tasks motivates a generalized compliance model relating arbitrary endpoint wrenches to corresponding deflection twists.

While our present control strategy does compensate for some structural effects, it assumes a quasistatic boom and does not account for dynamics and vibrations. Model-based approaches (e.g., MPC \cite{takacs2014adaptive}) can be utilized to better account and adapt to inherent compliances for faster, stable control. 

The current system also utilizes task motion primitives derived from human demonstration. This approach is simple and effective, but lacks generalization to similar tasks. For a scalable approach, we may develop better robot/task representations through imitation learning \cite{hussein2017imitation} and deformable linear object modeling \cite{saha2007manipulation} respectively. 


\subsection{Implications for Lunar Outfitting}

Looking further ahead, we recognize that cable routing represents just one of many outfitting operations required for future lunar infrastructure. Other tasks such as sensor placement, material transport, and connector mating share similar requirements: long reach, low interaction force, and large workspace access in constrained or elevated locations. Our approach offers a promising solution for these tasks.

Additionally, future lunar payloads (e.g. CLPS) will need to account for specific environmental factors such as reduced gravity, extreme temperatures, and abrasive regolith. Partial or microgravity environments reduce boom deflection but also weaken stabilizing forces, making the system  prone to inertia-driven oscillation. Dust mitigation and other environmental resilience considerations will also shape future design choices for boom deployment mechanisms and electronics.

While this work focuses on characterizing a single arm, we envision manipulators deployed from legged or wheeled mobile platforms on the lunar surface. Stability considerations will require future work in whole-body coordination \cite{khatib2008unified} and stance planning \cite{morton2024reachbot}. Recent work combining a quadruped with a robotic arm \cite{portela2024whole, magubane2025penn} highlights promising directions for coordinated manipulation and mobility. In addition, integrating multiple extendable arms could allow for bracing strategies \cite{book1985bracing}, bimanual manipulation, or even self-locomotion \cite{chen2024locomotion}. This would increase the stability, dexterity, and high-force interaction capability for a robot system.

\section{Conclusion}

This work presents a deployable, long-reach robotic manipulator capable of performing semi-autonomous cable routing tasks in large workspaces. This is enabled by a compact hardware design combining a deployable boom with a dexterous wrist, visual servoing, and a speed-aware control strategy informed by error modeling.

Experiments showed that our error model allowed reliable task execution by modulating task speed based on robot configuration. The system successfully routed cables across randomly distributed hooks, validating our approach. Looking ahead, we hope to increase system autonomy, incorporate dynamic modeling, and integrate our arm with mobile platforms for future lunar construction and outfitting tasks.


\section*{Acknowledgment}
\emph{Special Thanks:} The Stanford Robotics Center (SRC), including Matt Van Cleave, Eiko Rutherford, Zen Yaskawa, Steve Cousins, and Oussama Khatib.

\emph{Funding:} Stanley Wang and Venny Kojouharov were supported by the NSF Graduate Research Fellowship (with additional support for Venny from Knight-Hennessy Scholars), and Daniel Morton was supported by a NASA Space Technology Graduate Research Opportunity.

\emph{AI Use:} \cref{fig:concept} utilized DALL-E (OpenAI) for generation of the background image and rover asset. These were then composited with CAD renders of our manipulator arm.

\bibliographystyle{IEEEtran}
\bibliography{citations}


\end{document}